\documentclass{article}

\usepackage{PRIMEarxiv}

\usepackage[utf8]{inputenc} 
\usepackage[T1]{fontenc}    
\usepackage{hyperref}       
\usepackage{url}            
\usepackage{booktabs}       
\usepackage{amsfonts}       
\usepackage{nicefrac}       
\usepackage{microtype}      
\usepackage{lipsum}
\usepackage{fancyhdr}       
\usepackage{graphicx}       
\graphicspath{{media/}}     

\usepackage{amssymb}
\usepackage{amsmath}

\usepackage{mathtools}
\usepackage[belowskip=-5pt,aboveskip=2pt]{caption}

\usepackage{multirow}
\usepackage[table,xcdraw]{xcolor}

\title{A Novel Multi-Objective Velocity-Free Boolean Particle Swarm Optimization 
}

\author{
  Wei Quan \\
  Dept. of Computer Science \\
  University College London \\
  London, UK\\
  \texttt{erinquan0521@gmail.com} \\
   \And
  Denise Gorse \\
  Dept. of Computer Science \\
  University College London \\
  London, UK\\
  \texttt{d.gorse@cs.ucl.ac.uk} \\
}

\begin{document}
\maketitle

\begin{abstract}
This paper extends boolean particle swarm optimization to a multi-objective setting, to our knowledge for the first time in the literature. Our proposed new boolean algorithm, MBOnvPSO, is notably simplified by the omission of a velocity update rule and has enhanced exploration ability due to the inclusion of a “noise” term in the position update rule that prevents particles being trapped in local optima. Our algorithm additionally makes use of an external archive to store non-dominated solutions and implements crowding distance to encourage solution diversity. In benchmark tests, MBOnvPSO produced high quality Pareto fronts, when compared to benchmarked alternatives, for all of the multi-objective test functions considered, with competitive performance in search spaces with up to 600 discrete dimensions.
\end{abstract}

\keywords{Binary PSO \and Boolean PSO \and Multi-objective optimization \and Velocity-free}

\section{Introduction}
There are many real-world discrete optimization problems in both single and multi-objective settings, for example in electrical engineering \cite{coello1} \cite{zaharis} and feature selection \cite{nguyen}, and therefore a need for efficient and versatile algorithms that can solve such problems. Particle swarm optimization (PSO) \cite{kennedy1} and genetic algorithms (GAs) are both biologically inspired, effective algorithms, with discrete single and multi-objective variants; however, PSO has been shown to be less computationally demanding than GAs \cite{lee}. In this work we propose MBOnvPSO, a new, boolean PSO for application to multi-objective optimization problems such as feature selection \cite{nguyen}, a boolean (natively binary) approach being both efficient (as noted in \cite{marandi}) and of potential value in facilitating a hardware implementation. To our knowledge there have been no previous presentations of multi-objective boolean PSOs. This will be the primary novel contribution of our work.

Our second contribution will be to add to the evidence \cite{nguyen} \cite{shen1} that PSOs operating in a discrete search space may find good solutions more easily without a traditional velocity update rule, particularly for multi-objective problems, as evidenced in \cite{wang}. Velocity-free algorithms appear superior in this sense. Our MBOnvPSO algorithm, in which “nv” denotes “no velocity”, was inspired by the velocity-free binary PSO of \cite{shen1}, though with substantial differences (in addition to being boolean rather than binary) relating to how both integral and additional exploration are carried out. As will be seen later, MBOnvPSO has proved very effective when compared to the benchmarked alternatives, including a multi-objective variant of \cite{shen1}.

\section{Background and related work}
\label{sec:sec2}

\subsection{Continuous, binary, and boolean PSO}

PSO was first introduced in Kennedy and Eberhart \cite{kennedy1}. It is a nature-inspired population-based search algorithm modelled after the flocking behavior of birds and other animals. The PSO optimization process, for the $i$th particle, is guided by cognitive (via a \emph{personal best} position vector, $\textbf{p}_i$ and social (via a \emph{global best}, $\textbf{g}$) influences, and defined, for an $N$-dimensional search space, by the velocity and position update rules 

\begin{equation} \label{eq:eq1}
\textbf{v}_i^{t+1} = w \textbf{v}_i^t + c_1 r_1 ( \textbf{p}_i^t - \textbf{x}_i^t )+ c_2 r_2 ( \textbf{g}^t - \textbf{x}_i^t ),
\end{equation}

\begin{equation} \label{eq:eq2}
\textbf{x}_i^{t+1} =  \textbf{x}_i^t + \textbf{v}_i^{t+1},
\end{equation}

in which $\textbf{p}_i^t$  is the best parameter position found at time $t$ by particle $i$, $\textbf{g}^t$  the best position found by any particle, $c_1$  and $c_2$ are positive constants most often set to 2.0 \cite{wang}, $r_1$  and $r_2$  are random numbers $\in [0,1]$, and $w$ is a usually iteration-decreasing “inertia weight” that controls the extent to which new search directions are pursued. It is also usually recommended to clip the velocity to a maximum magnitude $V_{max}$  so that the search remains within useful bounds.

The update equations (\ref{eq:eq1}), (\ref{eq:eq2})
assume a continuous search space. However, Kennedy and Eberhart also proposed a binary PSO (BPSO) \cite{kennedy2}, in which to generate an updated binary position, velocities are transformed into $[0,1]$ by a sigmoid transfer function $sig()$, with 
$Prob(x_{ij} =1) = sig(v_{ij})$. The vast majority of later BPSOs have retained the concept of velocity, as well as the use of a transfer function to generate the velocity update. However, simpler BPSOs, without a velocity update rule, have proven to be equally, if not more, effective \cite{nguyen} \cite{shen1}, a central theme of this current work.

An alternative to the use of BPSOs in a discrete space is the boolean PSO of \cite{marandi}, which eliminates problems such as the velocity, in BPSO, not being associated with the likelihood of a change in position. Boolean PSO considers all variables to be binary and all arithmetic operators to be logical operators, with velocity and position update equations 

\begin{equation} \label{eq:eq3}
\textbf{v}_i^{t+1} = w . \textbf{v}_i^t + r_1 . ( \textbf{p}_i^t \oplus \textbf{x}_i^t ) + r_2 . ( \textbf{g}^t \oplus \textbf{x}_i^t ),
\end{equation}

\begin{equation} \label{eq:eq4}
\textbf{x}_i^{t+1} =  \textbf{x}_i^t \oplus \textbf{v}_i^{t+1},
\end{equation}

where “.” is the AND operator, “+” the OR operator, "$\oplus$" the XOR operator (all operators being applied bitwise), and where $Prob(r_1 =1) = \rho_1$, $Prob(r_2 =1) = \rho_2$,
$Prob(w =1) = \omega$.
We note that $V_{max}$ is now defined as the maximum number of velocity bits allowed to flip from 0 to 1 or 1 to 0.

There have been a number of applications of boolean single-objective PSO in electrical engineering that demonstrate its simplicity and computational efficiency, such as its application to antenna design in \cite{marandi} and \cite{kuhirun}. Outside of electrical engineering, a single-objective boolean PSO was used in \cite{gunasundari} for feature selection, with encouraging results that suggest a multi-objective variant, such as we propose here, might be an effective way forward in this area.

\subsection{Multi-objective optimization}

Unlike single-objective problems, multi-objective problems have conflicting objectives \cite{tan}. It is therefore necessary to obtain a \emph{non-dominated solution set}, based on \emph{Pareto dominance}, because no single solution can be considered ideal with respect to all objectives. The non-dominated set of the entire solution space is termed the \emph{Pareto-optimal set}, and the boundary of this region is termed the \emph{Pareto-optimal front}. Multi-objective PSO (MOPSO) was first proposed in \cite{coello2}, using Pareto dominance during the fitness evaluation stage. MOPSO then stores the non-dominated solutions produced by each particle in an external global best repository, the repository helping each particle to select a global best guidance during the search process.

\section{Benchmarked algorithms} \label{sec: sec3}

\begin{table}[]
\caption{Summary of benchmarks} \label{tab: tab1}
\centering
\begin{tabular}{|l|l|l|}
\hline
\textbf{Ref.} & \textbf{Acronyms} & \textbf{Binary or Boolean} \\ \hline
{\cite{gorse}}          & MBOPSO            & Boolean                    \\ \hline
{\cite{kennedy2}}           & MBPSO             & Binary                     \\ \hline
{\cite{mirjalili}}          & MVPSO             & Binary                     \\ \hline
{\cite{nguyen}}           & MSBPSO            & Binary                     \\ \hline
{\cite{shen1}}           & MBnvPSO           & Binary                     \\ \hline
\end{tabular}
\end{table}

This section will briefly describe each of the algorithms used for benchmarking, summarized in Table \ref{tab: tab1}. We note that examples of binary MOPSOs are relatively few, and that examples of boolean MOPSOs are limited to the current work. Hence, some benchmarks will be first uses of the underlying single-objective PSO in a multi-objective setting.

\subsection{Boolean benchmark}

As previously noted, there have been no previous examples of a boolean multi-objective PSO in the literature. However, it is possible to extend a single-objective boolean algorithm to the multi-objective domain. For this purpose, we choose the algorithm proposed by one of the current authors in \cite{gorse}, which added a “noise” term to the velocity update rule of \cite{marandi}, to enhance exploration, especially effectively in high dimensional search spaces. The velocity update rule of \cite{gorse} (the position update rule being equation (\ref{eq:eq4}), as in \cite{marandi}) is 

\begin{equation} \label{eq:eq5}
\textbf{v}_i^{t+1} = b + \bar{b}. w . \textbf{v}_i^t + r_1 . ( \textbf{p}_i^t \oplus \textbf{x}_i^t ) + r_2 . ( \textbf{g}^t \oplus \textbf{x}_i^t ),
\end{equation}

where the overbar denotes bitwise negation, and where $Prob(b = 1) = \beta$.  $\beta \in [0,1]$  is an exploration parameter that permits a small portion of the position update rule to be free from the influence of current velocity, and of personal or global best position. Our multi-objective implementation will be termed \emph{MBOPSO}.

\subsection{Binary benchmarks}

The most common BPSO to be extended to multi-objective problems is the original proposal of \cite{kennedy2}, used, for example, in \cite{malik} to prune a decision tree classifier, and in \cite{desouza} to select test cases in software testing. Though there are many closely related variants of \cite{kennedy2} (for example, the slightly modified sigmoid of \cite{habib}), as the representative of this most “traditional” group of BPSOs we have chosen to benchmark a multi-objective version of \cite{kennedy2}, given this algorithm's long history of use for multi-objective problems; this first of our binary benchmarks will be referred to \emph{MBPSO}.

Moving from sigmoid-type to differently shaped transfer functions, our next benchmark choice is the binary PSO with a V-shaped transfer function (\emph{VPSO}) proposed in \cite{mirjalili}. VPSO is a successful single-objective binary PSO, with more than 650 citations; however, it has never to our knowledge been used in a multi-objective setting. VPSO replaces the sigmoid function with a V-shaped function defined as

\begin{equation} \label{eq:eq6}
T(x) = \left \vert (2/ \pi) \tan^{-1}((\pi / 2) x) \right \vert.
\end{equation}

VPSO also changes the position update rule by assigning the previous position as the new position when the updated velocity is less or equal to a random number and assigning the complement of the previous position otherwise. We refer to its multi-objective variant as \emph{MVPSO}.

\emph{Sticky binary PSO} (SBPSO), proposed in \cite{nguyen} and applied to feature selection problems in \cite{li}, represents a more substantial departure from the original BPSO of \cite{kennedy2}. We have chosen this algorithm as a benchmark due to its recency and intrinsic interest. SBPSO replaces the velocity update rule with a “flipping probability”, with the update rule for the $j$th bit of particle $i$'s position given by

\begin{equation} \label{eq:eq7}
Prob(x_{ij}^{t+1} \rightarrow \bar{x}_{ij}^t) =
i_m ( 1 - s_{ij}^t) + i_p \left \vert p_{ij}^t - x_{ij}^t  \right \vert +
i_g \left \vert g_j^t - x_{ij}^t  \right \vert,
\end{equation}

where $s_{ij}^t$  is the particle’s “stickiness” (equivalent to momentum in a traditional PSO), $i_m$  controls the degree to which the stickiness operates, and $i_p$  and $i_g$  control the particle’s tendency to follow its personal and global best positions. We refer to our multi-objective version of SBPSO, the first such implementation in the literature, as \emph{MSBPSO}.

The most radical binary PSO we benchmark here in a multi-objective setting is the proposal of \cite{shen1}. This algorithm is also the closest in form to our own. The novelty in the BPSO of \cite{shen1} is its simplicity and its complete removal of the velocity update rule, with its position update rule defined by

\begin{equation} \label{eq:eq8}
x_{ij}^{t+1} = \begin{cases}
x_{ij}^t & \text{if } 0 < v_{ij} \leq \alpha\\
p_{ij}^t & \text{if } \alpha < v_{ij} \leq (1+\alpha)/2\\
g_j^t & \text{if } (1+\alpha)/2 < v_{ij} \leq 1
\end{cases},
\end{equation}

where $\alpha$ is a probability $\in (0,1)$  that determines the amount of exploration carried out; when $\alpha$  is smaller, it can be seen from equations (\ref{eq:eq8}) that the particle is less likely to remain in its current position. Additionally, to help prevent the algorithm from being trapped in local minima, a potential problem in all non-trivial optimization problems, as an integral part of the algorithm 10\% of the particles are selected to search randomly instead of following equations (\ref{eq:eq8}). This velocity-free algorithm has been used successfully in many areas \cite{shen2} and was additionally modified and applied to multi-objective problems in \cite{wang}. This modification, however, took the form only of further exploration via a mutation-like term, which in our own experiments we did not find helpful. Hence, as our final binary benchmark algorithm, we have selected to implement a multi-objective variant of the original algorithm of \cite{shen1}. It will be referred to as \emph{MBnvPSO}, with the “nv” to emphasize its radical discarding of the velocity update rule.

\section{Methodology}

\subsection{The proposed multi-objective boolean no-velocity particle swarm optimization (MBOnvPSO)}

Our proposed new algorithm, \emph{multi-objective boolean no-velocity PSO} (MBOnvPSO), like \cite{shen1}, removes the velocity update (indicated by the “nv” in its acronym), but differs substantially from \cite{shen1} in its basic mechanisms and in the way it handles exploration, this latter point relating to both exploration integral to the algorithm (i.e., is a part of the position update rule) and also to additional exploration carried out in order to help avoid local minima.

The MBOnvPSO position update rule (there is, of course, no velocity update for the algorithm) is given by

\begin{equation} \label{eq:eq9}
\textbf{x}_i^{t+1} = b . \bar{\textbf{x}}_i^t + 
\bar{b} . [a_1 . \textbf{x}_i^t + \bar{a}_1 . (\bar{a}_2 . \textbf{p}_i^t + a_2 . \textbf{g}^t)],
\end{equation}

in which all operators are boolean, as in equations (\ref{eq:eq3})-(\ref{eq:eq5}), in which the overbar denotes bitwise negation, as in equation (\ref{eq:eq5}), and in which $a_1$, $a_2$, and $b$ are bits generated with probabilities $Prob(a_1 =1)= \alpha_1$, $Prob(a_2 =1)= \alpha_2$, and $Prob(b =1)= \beta$. Clearly, while MBOnvPSO shares its “no-velocity” concept with \cite{shen1}, it differs substantially in its mechanisms, as can be seen by comparing the form of equation (\ref{eq:eq9}) to that of equations (\ref{eq:eq8}).

The parameter $\alpha_2$ in equation (\ref{eq:eq9}) balances the roles of social (i.e., following the global best, with probability $\alpha_2$) and cognitive (i.e., following the personal best, with probability $1 - \alpha_2$) guidance. Since there is no reason to weight these contributions differently, we take $\alpha_2 = 0.5$. The parameters $\alpha_1$ and $\beta$, however, are critical exploration parameters. Ideally, such parameters would be chosen for each problem using a grid search. However, rigorous searches of parameter spaces were unfeasible due to the computation time required. Hence parameters were selected in part guided by those values used in the work of others, and in part after a result of preliminary experimentation.

$\alpha_1$ is the probability the particle stays at its current position, as opposed to exploring by following either cognitive or social guidance, and is equivalent to the $\alpha$ parameter in \cite{shen1}. In MBOnvPSO, $\alpha_1$ is chosen to linearly increase from 0.3 to 0.7, so that the chance of exploring is higher earlier in the search process, later decreasing, to encourage convergence. However, the $\alpha$ parameter of \cite{shen1}, both in the original presentation and in subsequent works based on it (for example \cite{shen3}), is set to linearly \emph{decrease} (from 0.5 to 0.33), resulting in the opposite, namely more exploration \emph{later} in the search process. This seems counterintuitive for an optimization algorithm and likely to damage convergence, and, indeed, in preliminary experiments, proved to give inferior Pareto fronts when compared to our choice of a linearly increasing $\alpha_1$.

$\beta$ is another exploration-related parameter, which governs the amount of additional exploration carried out, separate from the intrinsic processes of the PSO, so particles are less likely to be trapped in local minima. The means by which this is done in MBOnvPSO is related to \cite{gorse}: when the “noise bit” $b=1$, which occurs with a probability $\beta$, 5\% of particles in each dimension have this probability of flipping a bit to 0 or 1 without influence from their current personal or global best positions. This is an instance in which we chose parameters differently from the recommendation of a previous work; in \cite{shen1}, 10\% of particles are bit-flipped with a probability of 1.0, but the much higher level of random search in \cite{shen1} has a damaging effect for most of the test functions in 
Table \ref{tab:tab2},  
as will be seen in the Results.

\subsection{Implementation of multi-objective optimization}

\paragraph{Pareto archive.} We implement an archiving system based on that of \cite{coello2}, having both a \emph{global best repository} and \emph{personal best repository}. In the case of the global best repository, each particle’s updated position and fitness values in the swarm are checked and filtered against those of the rest of the particles in the swarm. All the dominated solutions in that iteration are removed, while the remaining non-dominated solutions are then compared to the solutions saved in the global best repository, again removing any dominated solutions. In the case of the personal best repository, each particle’s updated objective value is compared to its personal best flight experience in the personal best repository in each subsequent iteration after the first. If such an objective value is dominated by any values saved in its personal best history, then this objective value and position will not be saved in the archive. Any dominated solutions in the personal best repository will also be removed.
\paragraph{Crowding distance for personal best and global best selection.} A single personal best solution and a single global best solution are selected using a \emph{crowding distance strategy} \cite{raquel} for the position update process. The crowding distance strategy was used to increase the diversity of the solutions, so that the personal and global best values selected from the personal best and global best repositories always come from the least populated solution regions.

\subsection{Performance metrics}

The performance metrics used in this research are those typically used in the multi-objective optimization literature, namely \emph{hypervolume} (HV),

\begin{equation} \label{eq:eq10}
HV = \bigcup_{x_i \in X} V_x,
\end{equation}

where $X$ is the set of all solutions on the obtained Pareto front, and $V_x$ is the volume for the corresponding point in $X$ and the reference point, generational distance (GD),

\begin{equation} \label{eq:eq11}
GD = \frac{\sqrt{\sum _{i=1}^{|Q|} d_i^2}}{|Q|},
\end{equation}

where Q is the obtained the Pareto front, and $d_i$ is the Euclidean distance between the $i$th solution on the obtained Pareto front and the nearest solution on the true Pareto front, and, finally, the \emph{number of points on the Pareto front} (NoP), whose definition is self-explanatory. 
Given a set of solutions and their corresponding objective values, the HV is the volume generated by a reference point and the points on the approximated Pareto front. GD is a measure of the closeness of the approximated Pareto front to the true Pareto front and considers both the accuracy and the diversity of the approximated Pareto front; we note that the requirement to know the true Pareto front will constrain our choice of test problems. HV and GD were calculated using the PyMoo library developed by Blank and Deb \cite{blank}.

\subsection{Test problems}
         
\begin{table}[]
\caption{Test functions} \label{tab:tab2}
\centering
\begin{tabular}{|l|}
\hline
\textbf{Schaeffer Function N.1} \\ \hline
  Minimize $ \begin{cases} f_1(x) = x^2 \\ f_2(x) = (x-2)^2 \end{cases}$ \\ $-10 \leq x \leq 10$                              \\ \hline
\textbf{ZDT Function N.1}       \\ \hline
 Minimize 
$ \begin{cases} f_1(\textbf{x}) = x_1 \\ 
f_2(\textbf{x}) = g(\textbf{x}) \cdot h(f_1(\textbf{x}),g(\textbf{x}))\\ 
g(\textbf{x}) = 1 + \frac{9}{29} \sum _{k=2}^{n} x_k \\
h(f_1(\textbf{x}),g(\textbf{x})) = 1 - \sqrt{\frac{f_1(\textbf{x})}{g(\textbf{x})}} \end{cases}$
\\ $0 \leq x_k \leq 1$, $k=1,\ldots,n$, where $n=30$   \\ \hline
\textbf{ZDT Function N.2}       \\ \hline
  Minimize 
$ \begin{cases} f_1(\textbf{x}) = x_1 \\ 
f_2(\textbf{x}) = g(\textbf{x}) \cdot h(f_1(\textbf{x}),g(\textbf{x}))\\ 
g(\textbf{x}) = 1 + \frac{9}{n-1} \sum _{k=2}^{n} x_k \\
h(f_1(\textbf{x}),g(\textbf{x})) = 1 - (\frac{f_1(\textbf{x})}{g(\textbf{x})})^2 \end{cases}$ \\ 
$0 \leq x_k \leq 1$, $k=1,\ldots,n$, where $n=30$     \\ \hline
\textbf{ZDT Function N.3}       \\ \hline
  Minimize 
$ \begin{cases} f_1(\textbf{x}) = x_1 \\ 
f_2(\textbf{x}) = g(\textbf{x}) \cdot h(f_1(\textbf{x}),g(\textbf{x}))\\ 
g(\textbf{x}) = 1 + \frac{9}{n-1} \sum _{k=2}^{n} x_k \\
h(f_1(\textbf{x}),g(\textbf{x})) = 1 - \sqrt{\frac{f_1(\textbf{x})}{g(\textbf{x})}} 
- (\frac{f_1(\textbf{x})}{g(\textbf{x})}) \sin{(10 \pi f_1)} \end{cases}$ \\ 
$0 \leq x_k \leq 1$, $k=1,\ldots,n$, where $n=30$   \\ \hline
\textbf{ZDT Function N.4}       \\ \hline
 Minimize
 $ \begin{cases} f_1(\textbf{x}) = x_1 \\ 
f_2(\textbf{x}) = g(\textbf{x}) \cdot h(f_1(\textbf{x}),g(\textbf{x}))\\ 
g(\textbf{x}) = 1 + 10(n-1) + \sum _{k=2}^{n} ( x_k^2 - 10 \cos{(4 \pi x_k))} \\
h(f_1(\textbf{x}),g(\textbf{x})) = 1 - \sqrt{\frac{f_1(\textbf{x})}{g(\textbf{x})}} 
 \end{cases}$  \\
 $0 \leq x_1 \leq 1$; $-10 \leq x_k \leq 10$,  $k=2, \dots ,n$, where $n=10$\\ \hline
\textbf{ZDT Function N.6}       \\ \hline
 Minimize
 $ \begin{cases} f_1(\textbf{x}) = 1 - \text{e}^{(-4 x_1)} \text{sin}^6 (6 \pi x_1) \\ 
f_2(\textbf{x}) = g(\textbf{x}) \cdot h(f_1(\textbf{x}),g(\textbf{x}))\\ 
g(\textbf{x}) = 1 + 9 [\frac{\sum _{k=2}^{n} x_k}{9}]^{0.25}  \\
h(f_1(\textbf{x}),g(\textbf{x})) = 1 - (\frac{f_1(\textbf{x})}{g(\textbf{x})})^2 
 \end{cases}$  \\
 $0 \leq x_k \leq 1$,  $k=1, \dots ,n$, where $n=10$ \\ \hline
\end{tabular}
\end{table}

Test problems for multi-objective optimization, particularly ones for which the true Pareto fronts are known, are limited. The test functions 
in Table \ref{tab:tab2}, on the following page, 
are frequently used in work on multi-objective optimization and have also been used in this work. Note that in this table the $x_k$, with $i = 1, \ldots ,n$, are real-valued, i.e., from the original definitions of these multi-objective problems; in our implementations of the binary and boolean algorithms, each $x_k$ will become a $B$-dimensional bit string. The Schaeffer function N.1 is the easiest amongst those listed, as it only has one real dimension, and hence its solution space is that of 20-bit strings. Conversely, the ZDT function N.4 is the hardest, as it has $21^9$ different local Pareto-optimal fronts \cite{wang}.

\subsection{Experimental setup}

100 runs of the search process were carried out, in each case using 100 particles and 300 iterations. A bit string of length $B = 20$ was used to represent each real-valued variable in the test functions, such that the search is carried out in an $nB$-dimensional discrete (boolean) space. $B = 20$ was picked as being the most common choice in work using boolean PSO for single-objective function optimization, such as \cite{gorse}. 

Parameters governing the benchmark algorithms were set to values advised in the relevant papers. For MBPSO and MVPSO, $c_1$ and $c_2$ were set to be 2, while the inertia weight parameter decreased linearly from 0.9 to 0.4. For MSBPSO, $i_m = 0.25$, $i_b = 0.25$, $i_g = 0.5$, and the step parameter was set to 50. For MBnvPSO, $\alpha$ linearly decreased from 0.5 to 0.33. For our MBOnvPSO, the $\beta$ parameter was set at 0.5. 

\section{Results}

\begin{table}[]
\centering
\caption{Test results for all six algorithms} \label{tab:tab3}
\begin{tabular}{|c|c|llllll|}
\hline
                                                                                    &                                   & \multicolumn{6}{c|}{\textbf{Algorithm}}                                                                                                                                                                                                                                                                         \\ \cline{3-8} 
\multirow{-2}{*}{\textbf{\begin{tabular}[c]{@{}c@{}}Test \\ Function\end{tabular}}} & \multirow{-2}{*}{\textbf{Metric}} & \multicolumn{1}{c|}{\textit{\textbf{MBPSO}}} & \multicolumn{1}{c|}{\textit{\textbf{MVPSO}}} & \multicolumn{1}{c|}{\textit{\textbf{MOBPSO}}} & \multicolumn{1}{c|}{\textit{\textbf{MSBPSO}}} & \multicolumn{1}{c|}{\textit{\textbf{MBnvPSO}}}                  & \multicolumn{1}{c|}{\textit{\textbf{MBOnvPSO}}} \\ \hline
                                                                                    & \textbf{GD}                       & \multicolumn{1}{l|}{\textbf{0.003±0.00}}     & \multicolumn{1}{l|}{\textbf{0.003±0.00}}     & \multicolumn{1}{l|}{\textbf{0.003±0.00}}      & \multicolumn{1}{l|}{\textbf{0.003±0.00}}      & \multicolumn{1}{l|}{\textbf{0.003± 0.00}}                       & \textbf{0.003±0.00}                             \\ \cline{2-8} 
                                                                                    & \textbf{HV}                       & \multicolumn{1}{l|}{\textbf{0.887±0.00}}     & \multicolumn{1}{l|}{\textbf{0.884±0.00}}     & \multicolumn{1}{l|}{\textbf{0.887±0.00}}      & \multicolumn{1}{l|}{\textbf{0.887±0.00}}      & \multicolumn{1}{l|}{\textbf{0.887±0.00}}                         & \textbf{0.887±0.00}                             \\ \cline{2-8} 
\multirow{-3}{*}{\textbf{Schaeffer}}                                                & \textbf{NoP}                      & \multicolumn{1}{l|}{5294±224}                & \multicolumn{1}{l|}{1095±36}                 & \multicolumn{1}{l|}{6588±353}                 & \multicolumn{1}{l|}{6551±179}                 & \multicolumn{1}{l|}{6691±243}                                   & \cellcolor[HTML]{C0C0C0}\textbf{6999±288}       \\ \hline
                                                                                    & \textbf{GD}                       & \multicolumn{1}{l|}{0.254±0.08}              & \multicolumn{1}{l|}{0.229± 0.07}             & \multicolumn{1}{l|}{0.175±0.06}               & \multicolumn{1}{l|}{0.098±0.03}               & \multicolumn{1}{l|}{0.006±0.00}                                 & \cellcolor[HTML]{C0C0C0}\textbf{0.005±0.00}     \\ \cline{2-8} 
                                                                                    & \textbf{HV}                       & \multicolumn{1}{l|}{0.389±0.09}              & \multicolumn{1}{l|}{0.464±0.11}              & \multicolumn{1}{l|}{0.590±0.07}               & \multicolumn{1}{l|}{0.707±0.04}               & \multicolumn{1}{l|}{0.871±0.00}                                 & \cellcolor[HTML]{C0C0C0}\textbf{0.873±0.00}     \\ \cline{2-8} 
\multirow{-3}{*}{\textbf{ZDT1}}                                                     & \textbf{NoP}                      & \multicolumn{1}{l|}{6.08±1.9}                & \multicolumn{1}{l|}{8.65±3.3}                & \multicolumn{1}{l|}{17.80±5.0}                & \multicolumn{1}{l|}{25.35±4.3}                & \multicolumn{1}{l|}{\textbf{511.3±170}}                         & \textbf{566.7±204}                              \\ \hline
                                                                                    & \textbf{GD}                       & \multicolumn{1}{l|}{0.179±0.09}              & \multicolumn{1}{l|}{0.121±0.07}              & \multicolumn{1}{l|}{0.152±0.05}               & \multicolumn{1}{l|}{0.123± 0.04}              & \multicolumn{1}{l|}{\textbf{0.004±0.00}}                        & \textbf{0.004±0.00}                             \\ \cline{2-8} 
                                                                                    & \textbf{HV}                       & \multicolumn{1}{l|}{0.278±0.09}              & \multicolumn{1}{l|}{0.469±0.11}              & \multicolumn{1}{l|}{0.490±0.07}               & \multicolumn{1}{l|}{0.497±0.06}               & \multicolumn{1}{l|}{\textbf{0.769±0.01}}                        & \textbf{0.768±0.01}                             \\ \cline{2-8} 
\multirow{-3}{*}{\textbf{ZDT2}}                                                     & \textbf{NoP}                      & \multicolumn{1}{l|}{4.07±1.7}                & \multicolumn{1}{l|}{10.09±4.0}               & \multicolumn{1}{l|}{16.84±4.1}                & \multicolumn{1}{l|}{13.19±3.3}                & \multicolumn{1}{l|}{\textbf{808.0±370}}                         & 710.1±333                                       \\ \hline
                                                                                    & \textbf{GD}                       & \multicolumn{1}{l|}{0.216±0.09}              & \multicolumn{1}{l|}{0.334±0.14}              & \multicolumn{1}{l|}{0.147±0.05}               & \multicolumn{1}{l|}{0.067±0.03}               & \multicolumn{1}{l|}{0.007±0.00}                                 & \cellcolor[HTML]{C0C0C0}\textbf{0.006±0.00}     \\ \cline{2-8} 
                                                                                    & \textbf{HV}                       & \multicolumn{1}{l|}{0.394±0.08}              & \multicolumn{1}{l|}{0.207±0.10}              & \multicolumn{1}{l|}{0.522±0.07}               & \multicolumn{1}{l|}{0.668±0.05}               & \multicolumn{1}{l|}{0.852±0.01}                                 & \cellcolor[HTML]{C0C0C0}\textbf{0.858±0.01}     \\ \cline{2-8} 
\multirow{-3}{*}{\textbf{ZDT3}}                                                     & \textbf{NoP}                      & \multicolumn{1}{l|}{11.5±3.0}                & \multicolumn{1}{l|}{7.15±2.1}                & \multicolumn{1}{l|}{18.89±4.3}                & \multicolumn{1}{l|}{28.88±5.2}                & \multicolumn{1}{l|}{\textbf{114.7±39.0}}                        & \textbf{111.9±41}                               \\ \hline
                                                                                    & \textbf{GD}                       & \multicolumn{1}{l|}{59.9±6.8}                & \multicolumn{1}{l|}{71.03±7.3}               & \multicolumn{1}{l|}{20.99±13.3}               & \multicolumn{1}{l|}{23.00±6.8}                & \multicolumn{1}{l|}{\textbf{6.97±3.7}}                          & \textbf{7.20±5.1}                               \\ \cline{2-8} 
                                                                                    & \textbf{HV}                       & \multicolumn{1}{l|}{0.0±0.0}                 & \multicolumn{1}{l|}{0.0±0.0}                 & \multicolumn{1}{l|}{0.0±0.0}                  & \multicolumn{1}{l|}{0.0±0.0}                  & \multicolumn{1}{l|}{\textbf{0.032±0.17}}                        & \textbf{0.055±0.22}                             \\ \cline{2-8} 
\multirow{-3}{*}{\textbf{ZDT4}}                                                     & \textbf{NoP}                      & \multicolumn{1}{l|}{6.48±2.4}                & \multicolumn{1}{l|}{8.06±3.0}                & \multicolumn{1}{l|}{4.38±2.5}                 & \multicolumn{1}{l|}{4.25±1.6}                 & \multicolumn{1}{l|}{\cellcolor[HTML]{C0C0C0}\textbf{282.1±250}} & 168.7±168                                       \\ \hline
                                                                                    & \textbf{GD}                       & \multicolumn{1}{l|}{6.604±0.41}              & \multicolumn{1}{l|}{6.756±0.49}              & \multicolumn{1}{l|}{4.742±0.50}               & \multicolumn{1}{l|}{4.966±0.34}               & \multicolumn{1}{l|}{\textbf{0.457±0.44}}                        & \textbf{0.446±0.48}                             \\ \cline{2-8} 
                                                                                    & \textbf{HV}                       & \multicolumn{1}{l|}{0.017±0.04}              & \multicolumn{1}{l|}{0.015±0.04}              & \multicolumn{1}{l|}{0.016±0.05}               & \multicolumn{1}{l|}{0.007±0.04}               & \multicolumn{1}{l|}{\textbf{0.279±0.27}}                        & \textbf{0.305±0.28}                             \\ \cline{2-8} 
\multirow{-3}{*}{\textbf{ZDT6}}                                                     & \textbf{NoP}                      & \multicolumn{1}{l|}{15.06±3.2}               & \multicolumn{1}{l|}{14.23±3.4}               & \multicolumn{1}{l|}{12.52±3.7}                & \multicolumn{1}{l|}{14.58±3.8}                & \multicolumn{1}{l|}{\textbf{362.8±486}}                         & \textbf{245.6±295}                              \\ \hline
\end{tabular}
\end{table}

First, we consider examples of Pareto fronts from our experiments, before turning in the following subsection to a comparative analysis of the performance of the six considered algorithms, as summarized 
in Table \ref{tab:tab3}.

\subsection{Evaluation and comparison: Pareto fronts}

\begin{figure}[ht]
\centering 
\includegraphics[width=\textwidth]{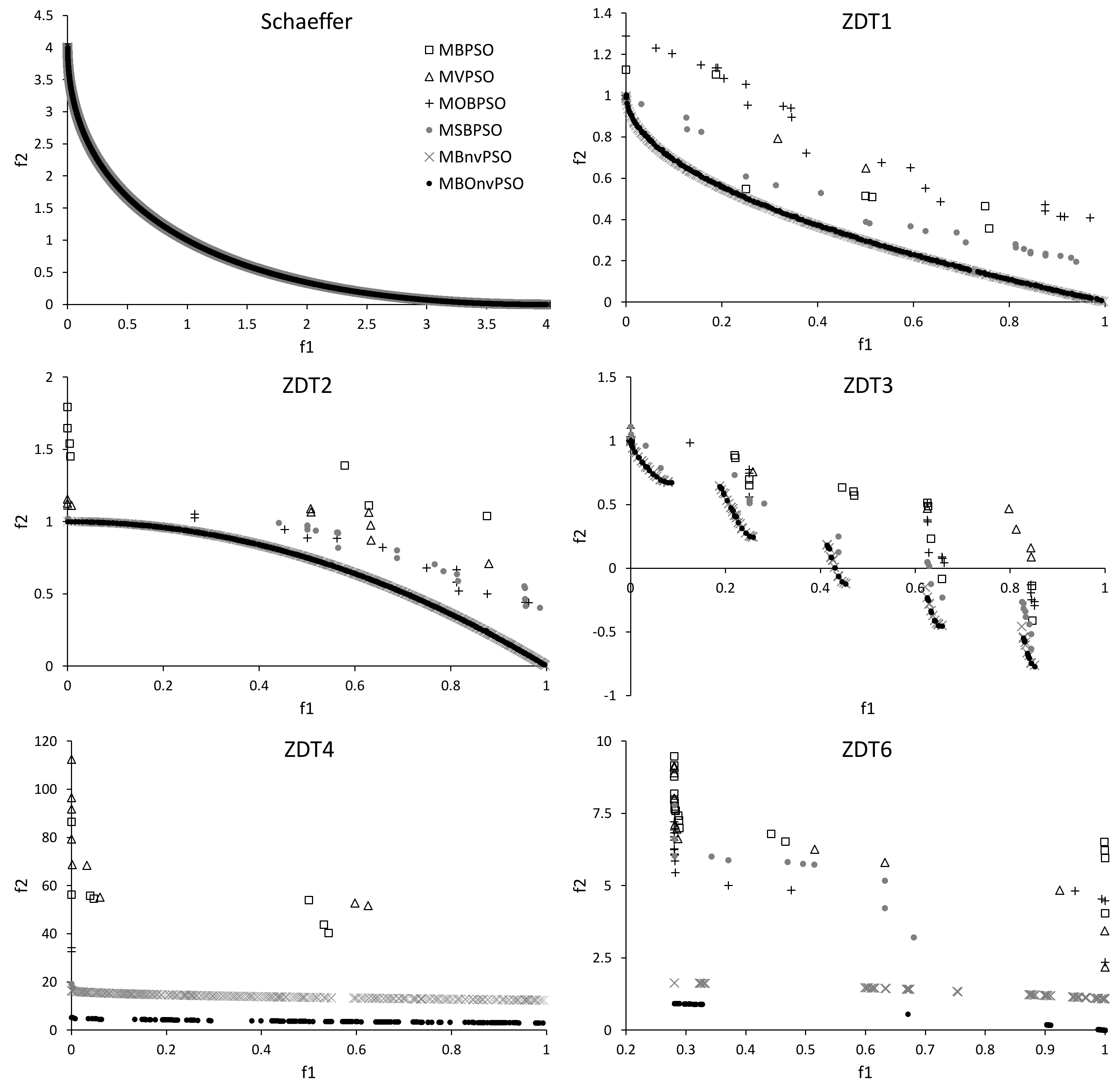}
\caption{Pareto fronts produced by MBOnvPSO and the five benchmark algorithms, for the Schaeffer function N.1 and the five ZDT functions.} 
\label{fig:fig1}
\end{figure}

The example Pareto fronts, shown in Figure \ref{fig:fig1} on p. 8, are selected based on the median GDs over the 100 runs of the experiments. 

In the case of the Schaeffer function N.1, all six algorithms have approximated the Pareto front easily, which is unsurprising given the simplicity of this problem, with only one real dimension in the search space. However, for all five ZDT functions, it can be seen that the two velocity-free algorithms, the proposed MBOnvPSO and MBnvPSO, are clearly and visibly better at approximating the Pareto fronts, with MBOnvPSO producing better fronts for the ZDT4 and ZDT6 functions. The worst Pareto fronts, in relation to both the number of discovered points and their distance from the true Pareto fronts, were produced by MBPSO and MVPSO, suggesting algorithms which utilize transfer functions in their implementation of velocity may be especially vulnerable to producing only a small number of low-quality trade-off solutions.

\subsection{Evaluation and comparison: results of Table \ref{tab:tab3}}

To further investigate the quality of the obtained Pareto fronts 
in Figure \ref{fig:fig1}
we measured the HV, GD, and NoP values of the fronts for all six algorithms. The results for our MBOnvPSO and the five benchmark algorithms are presented 
in Table 
and discussed below. Note that the table does not distinguish between performances which are not statistically significantly different; there are thus a number of “ties” 
in Table \ref{tab:tab3}. 
It does, however, distinguish between a win or tie at $0.001 < p \leq 0.05$ (bold font only) and a win at $p \leq 0.001$  (bold font and grey shading).

\subsubsection{Schaeffer function N.1}
The Schaeffer function N.1 is the easiest problem among the test functions because it has only one real-valued variable. Unsurprisingly, therefore, the difference in the quality of the Pareto fronts found by the six algorithms is small, with a statistically significantly different performance seen only in the case of the NoP metric, where our MBOnvPSO was the best-performing of the algorithms.

\subsubsection{ZDT functions}  
For each of the ZDT functions, it can be seen that one or both of the two velocity-free algorithms (our MBOnvPSO and the MBnvPSO of \cite{shen1}) in every case does statistically significantly better than those four algorithms that include a traditional concept of velocity (MBOPSO, MBPSO, MVPSO, MSBPSO). This difference is most marked for the ZDT4 function, acknowledged to be the most difficult of the set \cite{wang}, with the results produced by MBOnvPSO and MBnvPSO here drastically different from the other four algorithms. It is notable that the HV values for all of the algorithms that use a velocity concept are zero; this situation arises due to the highly inaccurate location of the front in these cases, together with the use of a standard definition of HV reference point. Of the two velocity-free algorithms, our MBOnvPSO has more outright wins at $p \leq 0.001$ than does MBnvPSO (four as opposed to one), and thus may be considered overall the best of the algorithms trialed. However, it is beaten twice by MBnvPSO on the NoP metric, once at $0.001 < p \leq 0.05$, in the case of ZDT2, and once at $p \leq 0.001$, in the case of ZDT4, for possible reasons that are discussed further below.

\subsubsection{Discussion of results}
Table \ref{tab:tab3}
evidences MBOnvPSO to be the overall best-performing algorithm, with statistically significantly better results than all five benchmarks five out of eighteen times, while its closest competitor, the binary velocity-free MBnvPSO, has statistically significantly better results than all other competitors only two out of eighteen times. It is noteworthy that the two best-performing algorithms are those that discard the concept of velocity.

Turning then, to a more detailed comparison of MBOnvPSO and MBnvPSO, one possible reason for the generally superior performance of MBOnvPSO lies in the way the additional search (beyond what is done within the position update rule) is performed. In order to prevent particles from being trapped in local minima, MBOnvPSO introduces randomness by allowing 5\% of particles a 50\% chance in each dimension to flip a bit randomly without influence from their current personal or global best positions, while in contrast MBnvPSO’s method forces 10\% of the particles to search randomly. MBOnvPSO's more moderate use of the additional random search may better preserve the information gained in the PSO search process.

We also note that MBnvPSO, following the construction of the single-objective algorithm on which it was based \cite{shen1}, caused the amount of search within the PSO update rule itself to increase with the number of iterations, rather than decrease, as would be more usually done to promote convergence. These choices would have caused MBnvPSO to utilize a level of search that would possibly inhibit convergence in the majority of cases. However, there may be situations in which the higher level of search in MBnvPSO is beneficial; the proposed MBOnvPSO is beaten by MBnvPSO, both times with respect to the NoP metric, for the functions ZDT2 and ZDT4, with this most marked ($p < 0.001$) in the case of ZDT4, which has many local minima.

\section{Conclusion}

This work has introduced MBOnvPSO, a novel boolean PSO algorithm for multi-objective optimization. In addition to its use of boolean operators in the position update process and an extra exploration capacity through the added “noise” term in the position update rule, the proposed MBOnvPSO notably removes the traditional PSO velocity update rule, as indicated by the “nv” (“no velocity”) in its acronym, this latter feature appearing to be of substantial importance in its success. 

In benchmarking tests, the proposed MBOnvPSO was compared to five other algorithms, including the multi-objective variant of a single-objective boolean PSO proposed by one of the current authors \cite{gorse}—this algorithm, MOBPSO, along with MBOnvPSO, being the first implementations of  a boolean multi-objective PSO in the literature—and a single-objective binary PSO that is also velocity-free \cite{shen1} (its multi-objective variant, as implemented by us, being termed here MBnvPSO).

The two velocity-free algorithms were, as noted, by far the best performing. It is also interesting to note that the third-best performing algorithm appeared to be MSBPSO (though this conclusion would likely require further runs to assure statistical significance); while MSBPSO does not discard velocity entirely, its implementation of the concept is non-standard. Based on the experimental results of this work, it thus appears the traditional PSO velocity update mechanism may not benefit the discrete search process for multi-objective optimization problems, and that this is especially evident with respect to the NoP metric: many more points are discovered by the velocity-free algorithms. 

Future work will take two directions. First, we will aim to improve MBOnvPSO's ability to handle problems such as ZDT4, with a large number of local minima, without using (as does MBnvPSO) an excess amount of undirected search that may be damaging to the performance in other cases. Second, as discussed in Section \ref{sec: sec3}, there are many instances of binary PSO application in feature selection in machine learning \cite{mafarja}\cite{ji}, and thus future work will also investigate the efficacy of the proposed MBOnvPSO in feature selection.

\bibliographystyle{unsrt}  
\bibliography{references}

\end{document}